\documentclass[letterpaper]{article} % DO NOT CHANGE THIS
\usepackage[submission]{aaai23}  % DO NOT CHANGE THIS
\usepackage{times}  % DO NOT CHANGE THIS
\usepackage{helvet}  % DO NOT CHANGE THIS
\usepackage{courier}  % DO NOT CHANGE THIS
\usepackage[hyphens]{url}  % DO NOT CHANGE THIS
\usepackage{graphicx} % DO NOT CHANGE THIS
\usepackage{natbib}  % DO NOT CHANGE THIS AND DO NOT ADD ANY OPTIONS TO IT
\usepackage{caption} % DO NOT CHANGE THIS AND DO NOT ADD ANY OPTIONS TO IT
\usepackage{algorithm}
\usepackage{algorithmic}

\usepackage{amsfonts}
\usepackage{multirow}
\usepackage{newfloat}
\usepackage{listings}
\usepackage{amsmath}
\DeclareCaptionStyle{ruled}{labelfont=normalfont,labelsep=colon,strut=off} % DO NOT CHANGE THIS
\floatstyle{ruled}
\newfloat{listing}{tb}{lst}{}
\floatname{listing}{Listing}
\title{An Automatic ICD Coding Network Using Partition-Based Label Attention}
\author{
    Daeseong Kim\textsuperscript{\rm 1}, 
    Haanju Yoo\textsuperscript{\rm 2},
    Sewon Kim\textsuperscript{\rm 2,*}
}
\affiliations{
    \textsuperscript{\rm 1}Korea Advanced Institute of Science \& Technology (KAIST), South Korea\\
    \textsuperscript{\rm 2}NAVER CLOVA AI Lab, South Korea\\
    eotjd0986@kaist.ac.kr, \{haanju.yoo, se1.kim\}@navercorp.com
}

\usepackage{bibentry}
\begin{document}

\maketitle

\begin{abstract}
International Classification of Diseases (ICD) is a global medical classification system which provides unique codes for diagnoses and procedures appropriate to a patient's clinical record. However, manual coding by human coders is expensive and error-prone. Automatic ICD coding has the potential to solve this problem. With the advancement of deep learning technologies, many deep learning-based methods for automatic ICD coding are being developed. In particular, a label attention mechanism is effective for multi-label classification, i.e., the ICD coding. It effectively obtains the label-specific representations from the input clinical records. However, because the existing label attention mechanism finds key tokens in the entire text at once, the important information dispersed in each paragraph may be omitted from the attention map. To overcome this, we propose a novel neural network architecture composed of two parts of encoders and two kinds of label attention layers. The input text is segmentally encoded in the former encoder and integrated by the follower. Then, the conventional and partition-based label attention mechanisms extract important global and local feature representations. Our classifier effectively integrates them to enhance the ICD coding performance. We verified the proposed method using the MIMIC-III, a benchmark dataset of the ICD coding. Our results show that our network improves the ICD coding performance based on the partition-based mechanism.
\end{abstract}

\begin{figure}[t!]
\centering
\includegraphics[width=1\linewidth]{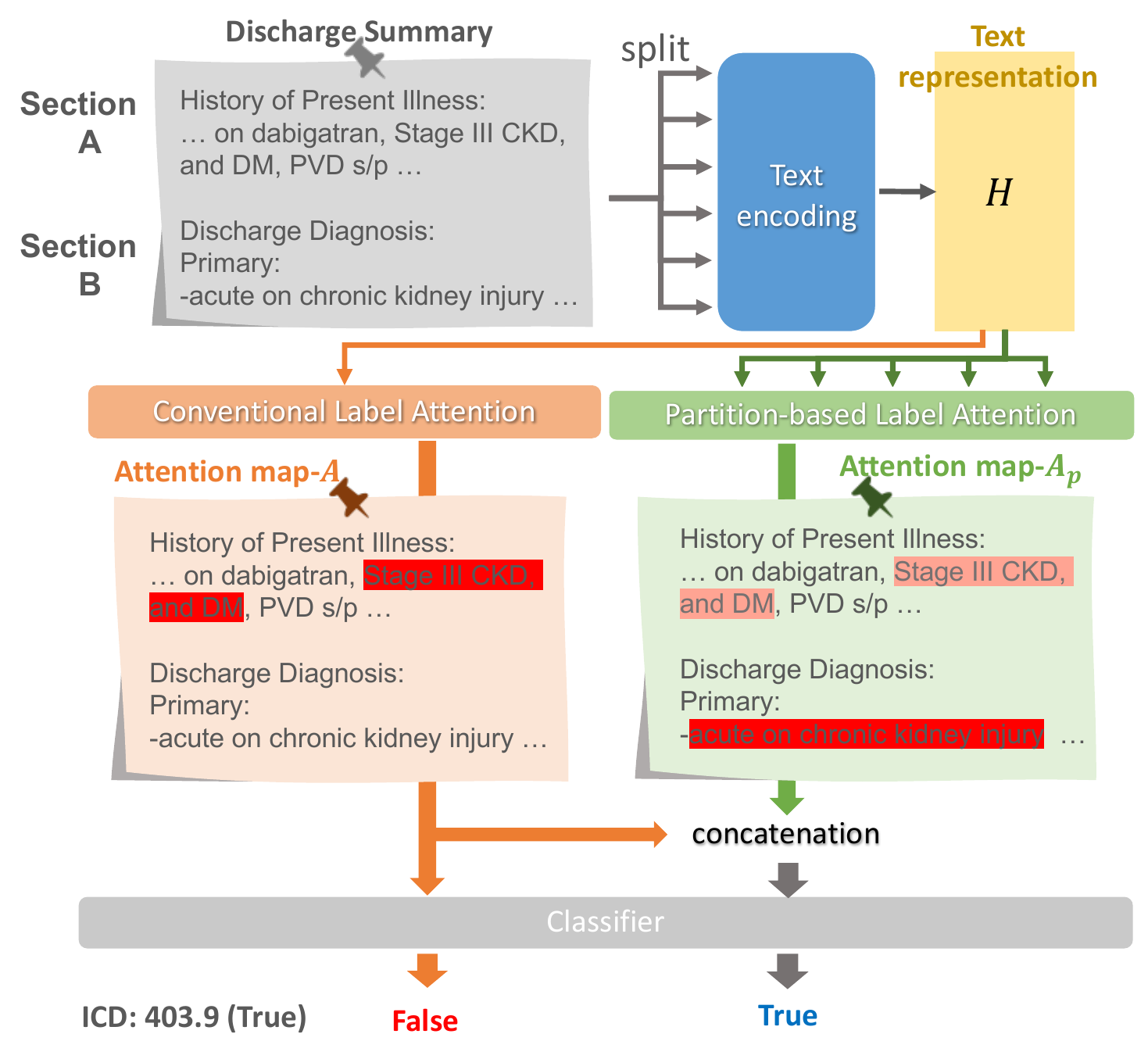}
\caption{The framework of PAAT. The input discharge summary is divided and passed to the encoding part. After the encoding, the conventional and partition-based attention mechanisms are applied to the text representation. The conventional attention mechanism captures the important tokens and the partition-based attention mechanism captures the informative dispersed tokens which the conventional label attention missed.}
\label{Fig.1}
\end{figure}

\section{Introduction}
International Classification of Disease (ICD) is a globally used medical classification system which provides unique codes for diagnoses and procedures performed during patient visits \cite{shull2019digital,ji2022unified}. ICD coding is the process of allocating appropriate ICD codes to a patient's clinical record. The allocated ICD codes are used for various medical and research purposes including epidemiological studies and service billing. The ICD coding process is typically performed manually by clinical coders, which makes it expensive and error-prone \cite{o2005measuring,adams2002addressing}. Automatic ICD coding potentially improves the quality of medical and related services by reducing human resource and errors.

Machine learning-based classification algorithms are representative methods for automatic ICD coding. Traditionally, automatic ICD coding systems extract features based on rules \cite{farkas2008automatic} or relying on expertise \cite{medori2010machine}. And they assign the ICD codes based on those features using mathematical regression algorithms such as the support vector machine (SVM) \cite{ferrao2013using} and Bayesian regression \cite{ji2022unified,lita2008large}. 

Recently, deep learning technologies achieved good performance in various tasks. Similarly in the automatic ICD coding, the introduction of deep learning-based networks has resulted in significant performance improvements. However, the current models still have difficulties in accurately extracting important information from the clinical notes. In clinical notes, the results of all examinations and interviews that patients experienced are written by item or chronologically. It may consist of one or several paragraphs rather than a single sentence, so important information for a final diagnosis is dispersed and hard to be grasped at once.

The label attention mechanism \cite{vu2021label} is an extended architecture from a structured self-attention \cite{lin2017structured}. It is designed to be suitable for multi-code assignments. Therefore, various studies applied the label attention mechanism to the automatic ICD coding \cite{sun2021multitask,biswas2021transicd,liu2022hierarchical}. However, because the label attention mechanism selects core tokens through a single softmax layer for the entire text, there is a risk that important tokens will be ignored in clinical notes where important information may be dispersed in long-length text. Dong et al. proposed a continuous label attention at the word-level and the sentence-level to learn the association between the latent representations and the corresponding labels \cite{dong2021explainable}. Inspired by the hierarchical label attention mechanism used in Dong et al., we propose an automatic ICD coding network using \textbf{PA}rtition-based label \textbf{AT}tention (PAAT). PAAT model consists of three parts: an encoding part, an attention part, and a classifier (see Fig. \ref{Fig.1}). The encoding part of the PAAT model segmentally encodes the input text considering the characteristic of clinical notes which were recorded separately for each topic. Then, the conventional and partition-based attention mechanisms are used to obtain globally and locally meaningful latent representations from the encoded feature, respectively. Finally, the classifier assigns the ICD codes based on both representations. The main contributions of our study are as follows:
\begin{itemize}
\item Our PAAT model captures locally scattered but important information. To this end, the PAAT model obtains an optimal text representation from the input text using the partition-based encoding. After that, it obtains global latent representations and local latent representations for each region using the conventional and partition-based label attention mechanisms, respectively. Finally, our local latent representations are integrated based on the importance of each region.
\item Our encoder, composed of Clinical-Longformer \cite{li2022clinical} and a bidirectional long short-term memory (bi-LSTM) layer, can be applied to long texts exceeding the maximum allowable input length of Longformer-based models, i.e., 4096 tokens, by segmentally encoding the input text.
\item We verified the performance of PAAT using MIMIC-III \cite{johnson2016mimic}, an anonymized real-world electronic medical records (EMR), which is widely used for EMR relevant studies including automatic ICD coding. In our experiments, PAAT classified the ICD codes from the discharge summaries and showed strong performance compared to existing methods.
\end{itemize}

\begin{figure*}[!t]
\centering
\includegraphics[width=1.7\columnwidth]{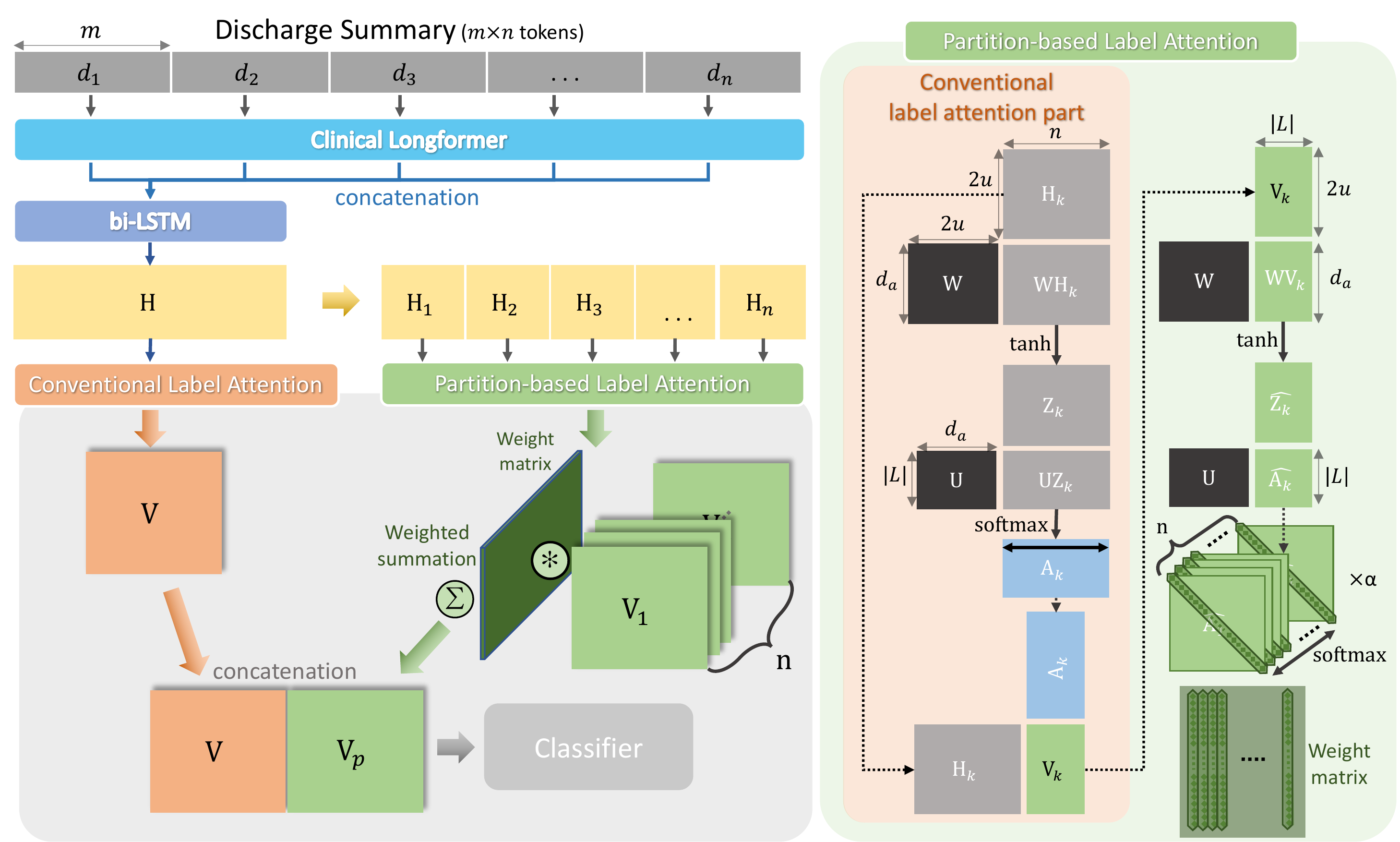}
\caption{An overview of the PAAT model. The text is segmentally encoded by the Clinical-Longformer and integrated by the bi-LSTM layer. The conventional attention mechanism is applied to the entire text representation, and the partition-based label attention mechanism is applied to the divided text representations. The final decision is based on the outputs of both label attention mechanisms.}
\label{Fig.2}
\end{figure*}

\section{Related Works}
Automatic ICD coding from the discharge summary was first tried in the 1990s. Larkey and Croft carried out automatic ICD coding through an ensemble of three classifiers based on K-nearest neighbor, relevance feedback, and Bayesian independence, respectively \cite{larkey1996combining}. De Lima et al. and Perotte et al. constructed ICD code classifiers that reflect the hierarchical structure of the ICD codes using SVM and cosine similarity between the discharge summary and the ICD code description, respectively \cite{de1998hierarchical,perotte2014diagnosis}.

Based on the success of deep learning technologies, many of automatic ICD coding studies have been applied deep learning algorithms and achieved significant performance improvement. The method proposed by Shi et al. is one of the earliest methods of applying recurrent neural networks to automatic ICD coding \cite{shi2017towards}. MultiResCNN \cite{li2020icd} uses multi-channel convolutional layers of different kernel sizes to capture various text patterns. It combines residual learning with multi-channel convolutional layers to improve learning efficiency.

The application of the attention mechanism improved the performance of ICD coding. Mullenbach et al. proposed a CAML architecture \cite{mullenbach2018explainable} for ICD coding. CAML encodes the input text using convolutional neural networks (CNN) and applies the attention mechanism to capture highly-predictive words for each label. MSATT-KG \cite{xie2019ehr} proposed by Xie et al. applied the densely connected CNN for text encoding to produce variable $n$-gram features. Then, a multi-scale feature attention is applied to capture the most informative $n$-gram features. Vu et al. applied a hierarchical decoder considering the hierarchical structure of ICD codes. It classifies the input text by using the label attention mechanism for high-level codes and then classifies low-level codes using the results of high-level codes \cite{vu2021label}.

Transformer \cite{vaswani2017attention}, which showed good performance of text encoding in various fields, also contributed to the improvement of ICD coding. TransICD \cite{biswas2021transicd} uses the transformer encoder for encoding discharge summaries to enhance the ICD coding performance. Ji et al. proposed BERT-hier \cite{ji2021does} which splits the long text into multiple chunks for applying BERT \cite{devlin2018bert} which is a self-supervised model pretrained with large unlabeled data using the transformer.

Some studies tried to improve the ICD coding performance by using the associations between the ICD codes. Sun et al. proposed an encoder that can refine text representations using down-sampling and up-sampling based on a convolution layer. And they jointly train the encoder with the ICD and CCS coding tasks using the label attention mechanism to improve the ICD coding performance \cite{sun2021multitask}. Zhou et al. proposed ISD \cite{zhou2021automatic} using the transformer decoder structure. ISD introduces a shared representation extraction mechanism based on both clinical notes and annotated codes, and a self-distillation learning mechanism to solve the long tail and noisy text problems of clinical records. While these works use the characteristics of the ICD code, they do not take into account the characteristics of the discharge summary.

A hierarchical approach was also attempted to improve the efficiency of the attention mechanism. Dong et al. applied a hierarchical attention mechanism that sequentially uses word-level and sentence-level attention \cite{dong2021explainable}. Liu et al. developed ClinicalPlusXLNet \cite{liu2022hierarchical}, which improved encoding performance for clinical notes by pretraining XLNet \cite{yang2019xlnet} using data from the MIMIC-III. And they also proposed HiLAT \cite{liu2022hierarchical} for the automatic ICD coding. It divided the input text into multiple segments because of the length limit of input tokens and encoded them based on ClinicalPlusXLNet. Afterward, the token-level attention for each segment is applied. Then, the obtained label-specific representations from segments are gathered along the labels, and segment-level label attentions were performed.

Unlike these studies, The PAAT model segmentally encodes the text based on the transformer considering the characteristic of the discharge summary. In addition, It obtains the global and local latent representations by applying both the conventional and partition-based label attention layers. Therefore, it efficiently captures the dispersed important information of the discharge summary without missing.

\section{Methods}
In this section, we introduce the PAAT model that assigns ICD codes to discharge summaries. The entire architecture of the PAAT model is described in Fig. \ref{Fig.2}. The PAAT model consists of the encoding part, the attention part composed of the conventional and partition-based attention layers, and the classifier. The input text is divided into $n$ segments and encoded separately with the Clinical-Longformer. Then, $n$ encoded text representations are integrated by the bi-LSTM layer. Then, two label attention mechanism are applied. First, the conventional label attention mechanism obtains the conventional label-specific matrix from the text representation. Second, the partition-based label attention mechanism divides the text representation again, and obtains the label specific matrix from each segment. Then, it integrates the matrices for segments to generate partition-based label-specific matrix. Finally, the classifier detects the ICD codes based on both the conventional and partition-based label-specific matrices.

\subsection{Encoding part of PAAT}
The pretrained transformers (e.g., BERT \cite{devlin2018bert}) and bi-LSTM have been used together in many studies as encoders to consider both global and sequential features \cite{shao2019transformer,vlad2019sentence,lee2019ncuee}. However, the BERT model has a short limit length of 512 for input tokens, which makes it unsuitable for paragraph-level text embedding. In addition, because it is trained with general domain text such as that from Wikipedia, it is hard to cover clinical expertise contained in the discharge summary \cite{lee2020biobert}. Instead of BERT, we used the Clinical-Longformer model \cite{li2022clinical}, which is trained with biomedical text and can embed long text up to 4096 tokens, with a bi-LSTM layer as the text encoder. 

Clinical notes contain the results of interviews and examinations, which are recorded according to topics such as `discharge diagnosis', `admission medications', and `past medical history'. Therefore, it is inefficient to obtain a text representation at once in multiple sections that have little relation to each other. We simplify the idea used in the work proposed by Pappagari et al. \cite{pappagari2019hierarchical} to segment the input text and encode them separately using the Clinical-Longformer. The text representation obtained for each segment is expected to be optimal for the corresponding segment. Afterward, an integrated latent representation is obtained from the text representations of all segments through the bi-LSTM layer.

\subsection{Label Attention Mechanism}
The label attention mechanism is a technique that simultaneously extracts label-specific information for all labels from the input text. Therefore, it can be applied to multi-label classification tasks such as ICD coding \cite{vu2021label,dong2021explainable,sun2021multitask,biswas2021transicd,liu2022hierarchical}. The label attention mechanism is expressed by the following equations:
\begin{equation}
Z=\tanh(WH),
\end{equation}
\begin{equation}
A=\text{softmax}(UZ),
\end{equation}
\begin{equation}
V=HA^\top,
\end{equation}
where $H \in \mathbb{R}^{2u \times N}$ indicates the hidden space feature obtained by the encoding part, $2u$ denotes the feature dimension of the bi-LSTM layer, and $N$ denotes the number of tokens in the input text. $W \in \mathbb{R}^{d_a \times 2u}$ and $U \in \mathbb{R}^{L \times d_a}$ are the trainable weight matrices, and $d_a$ and $L$ denote the depth of hidden space and the number of labels respectively. $A \in \mathbb{R}^{L \times N}$ which is obtained from $H$, $W$, and $U$ represents the attention weight matrix for the labels corresponding to the input tokens. Therefore, the $l^{th}$ column of a feature matrix $V \in \mathbb{R}^{2u \times L}$ represents the information about $l^{th}$ label. In other words, the label attention mechanism can obtain $V$, whose feature dimension is $2u$ corresponding to $L$ labels, from $N$ encoded tokens.

\subsection{Partition-based Label Attention Mechanism}
The label attention mechanism efficiently obtains label-specific features based on the matrix multiplication. However, it only pays attention to a relatively small number of tokens through a single softmax layer. As a result, some important information spread across the entire text may be missed. To overcome this, we propose a partition-based label attention mechanism. The partition-based label attention mechanism divides the text representation obtained from the encoder and generates label-specific features for each segment. Then it performs a weighted summation of the features to obtain a combined label specific feature matrix which covers important features dispersed across the entire text. This partition-based label attention mechanism is expressed by the following equations:
\begin{equation}
Z_k=\tanh(W H_k),
\end{equation}
\begin{equation}
A_k=\text{softmax}(U Z_k),
\end{equation}
\begin{equation}
V_k=H_k (A_k)^T,
\end{equation}
where $k\in\{1,2,...,n\}$ denotes the $k^{th}$ segment, and $n\in\mathbb{N}$ denotes the number of segments. This process is the result of the label attention mechanism on each segmented text representation. $(V_k)_l$ represents the information about $l^{th}$ label in the $k^{th}$ segment. Because $UZ_k$ acts as a quantitative indicator of the association between each input token and the label, $A_k$ acts as the attention weight of $H_k$ for each label . 

The weighted sum of feature matrices obtained for each segment is expressed as follows.
\begin{equation}
\widehat{Z_k}=\tanh(W V_k)
\end{equation}
\begin{equation}
\widehat{A_k}=U\widehat{Z_k}
\end{equation}
\begin{equation}
T_k=\text{diag}(\widehat{A_k})\times\alpha,
\end{equation}
\begin{equation}
M_l=\text{softmax}(\{(T_1)_l, (T_2)_l, ... (T_n)_l\}),
\end{equation}
\begin{equation}
(V_p)_l = \displaystyle\sum_{k=1} ^n (M_l)_k (V_k)_l,
\end{equation}
where $\alpha$ is a constant for smoothing the softmax transform in equation (10), and $(V_p)_l$ denotes the $l^{th}$ column of $V_p$. The equation (8) is an intermediate process in (5), which indicates a quantitative indicator of the association between the input representation and the labels. At this time, because $(V_k)_l$ represents the information for the $l^{th}$ label in the $k^{th}$ segment, as a result, $\widehat{A_k}$ represents the quantitative indicator of the association between the representation for each label in the $k^{th}$ input segment and each label. Therefore, $T_k$, the diagonal component of $\widehat{A_k}$, indicates the quantitative association between each label-specific latent representation of $k^{th}$ segment and the corresponding label. $M$ can be operated as a measure of association between each label and each segment by applying a softmax transform for $T$, and we obtained $(V_p)_l$ through the weighted sum of $(M_l)_k$ and $(V_k)_l$. 

Finally, the medical code is assigned by the classifier composed of a simple feed-forward network based on the label-specific matrices obtained by the conventional and partition-based label attention layers.

\begin{table}[!t]
\caption{Comparison of the baseline models and the PAAT model results on the MIMIC-III 50 datasets (in \%).}
\label{Table1}
\begin{tabular}{l|ccccc}
\hline
\multicolumn{1}{c|}{\multirow{3}{*}{Models}}                       & \multicolumn{5}{c}{MIMIC-III 50}                                                                                                                                                                                                                                                                                                \\
\multicolumn{1}{c|}{}                                              & \multicolumn{2}{c}{AUC}                                                                                                      & \multicolumn{2}{c}{F1}                                                                                                                & \multirow{2}{*}{P@8}                                     \\
\multicolumn{1}{c|}{}                                              & \multicolumn{1}{l}{macro}                                & \multicolumn{1}{l}{micro}                                         & \multicolumn{1}{l}{macro}                                         & \multicolumn{1}{l}{micro}                                         &                                                          \\ \hline
MSATT-KG                                                           & 91.4                                                     & 93.6                                                              & 63.8                                                              & 68.4                                                              & 64.4                                                     \\
MultiResCNN                                                        & 89.9                                                     & 92.8                                                              & 60.6                                                              & 67.0                                                              & 64.1                                                     \\
LAAT                                                               & 92.5                                                     & 94.6                                                              & 66.6                                                              & 71.5                                                              & 67.5                                                     \\
JointLAAT                                                          & 92.5                                                     & 94.6                                                              & 66.1                                                              & 71.6                                                              & 67.1                                                     \\
ISD                                                                & \textbf{93.5}                                            & 94.9                                                              & 67.9                                                              & 71.7                                                              & \textbf{68.2}                                            \\
MARN                                                               & 92.7                                                     & 94.7                                                              & 68.2                                                              & 71.8                                                              & 67.3                                                     \\ \hline
PAAT                                                               & \begin{tabular}[c]{@{}c@{}}92.8\\ $\pm$ 0.1\end{tabular} & \begin{tabular}[c]{@{}c@{}}94.8\\ $\pm$ 0.1\end{tabular}          & \textbf{\begin{tabular}[c]{@{}c@{}}68.5\\ $\pm$ 0.5\end{tabular}} & \textbf{\begin{tabular}[c]{@{}c@{}}73.0\\ $\pm$ 0.1\end{tabular}} & \begin{tabular}[c]{@{}c@{}}67.8\\ $\pm$ 0.2\end{tabular} \\
\begin{tabular}[c]{@{}l@{}}PAAT\\ +label \\ embedding\end{tabular} & \begin{tabular}[c]{@{}c@{}}92.8\\ $\pm$ 0.0\end{tabular} & \textbf{\begin{tabular}[c]{@{}c@{}}94.9\\ $\pm$ 0.1\end{tabular}} & \begin{tabular}[c]{@{}c@{}}68.3\\ $\pm$ 0.2\end{tabular}          & \begin{tabular}[c]{@{}c@{}}73.0\\ $\pm$ 0.1\end{tabular}          & \begin{tabular}[c]{@{}c@{}}68.0\\ $\pm$ 0.3\end{tabular} \\ \hline
HiLAT                                                              & 92.7                                                     & 95.0                                                              & 69.0                                                              & 73.5                                                              & 68.1                                                     \\
\begin{tabular}[c]{@{}l@{}}HiLAT\\ (reproduce)\end{tabular}        & \begin{tabular}[c]{@{}c@{}}92.9\\ $\pm$ 0.2\end{tabular} & \begin{tabular}[c]{@{}c@{}}95.0\\ $\pm$ 0.1\end{tabular}          & \begin{tabular}[c]{@{}c@{}}68.2\\ $\pm$ 0.6\end{tabular}          & \begin{tabular}[c]{@{}c@{}}73.2\\ $\pm$ 0.2\end{tabular}          & \begin{tabular}[c]{@{}c@{}}68.0\\ $\pm$ 0.1\end{tabular} \\ \hline
\end{tabular}
\end{table}

\section{Experiments}
\subsection{Datasets}
The MIMIC-III is a benchmark dataset that contains medical information of over 40,000 patients in the intensive care unit of Beth Israel Deaconess Medical Center from 2001 to 2012. We used discharge summaries and human annotated ICD-9 codes from the MIMIC-III dataset like previous studies of ICD coding. Our preprocessing process follows that of CAML and LAAT \cite{mullenbach2018explainable,vu2021label}. After the preprocessing, there are 52,722 discharge summaries and total of 8,907 unique codes in the MIMIC-III dataset. We conducted two experiments based on the MIMIC-III dataset. In the first experiment, 47,719 of the entire discharge summaries were used for training, 1,631 for validation, and 3,372 for testing. In the second experiment, 50 the most frequently occurring codes and 11,368 corresponding discharge summaries were used, of which 8,066 were used for training, 1,573 for validation, and 1,729 for test. All divisions were based on patient ID.

\subsection{Metrics and Settings}
For the same comparison with previous studies on ICD coding, we used the macro and micro areas under the receiver operating characteristic curve (AUC), macro and micro F1-scores, and precision at k (P@k) as evaluation metrics. We used the micro F1-score for validation on the evaluation set. Except for the results of the baseline methods, all our experimental results are the average of the results performed based on 5 different random seeds. 

We used the pretrained Clinical-Longformer from HuggingFace which is fixed during the learning process. The maximum number of input tokens is 8192 with partition, and 4096 without partition. The hidden size of the bi-LSTM layer, $2u$, was 1024 (512$\times$2), and $d_a$ and $\alpha$ were set to 512 and 0.8, respectively. The dropout rate of 0.3 was applied in the learning process for the bi-LSTM layer. We used an AdamW optimizer with a learning rate of 0.0015. The binary cross-entropy loss function was used for the training of our model. Our entire code was written based on the LAAT \cite{vu2021label}.

\begin{table}[t]
\caption{Comparison of the baseline models and the PAAT model results on the MIMIC-III Full datasets (in \%).}
\label{Table2}
\begin{tabular}{l|ccccc}
\hline
\multicolumn{1}{c|}{\multirow{3}{*}{Models}}                       & \multicolumn{5}{c}{MIMIC-III Full}                                                                                                                                                                                                                                                                                              \\
\multicolumn{1}{c|}{}                                              & \multicolumn{2}{c}{AUC}                                                                                                      & \multicolumn{2}{c}{F1}                                                                                                       & \multirow{2}{*}{P@5}                                              \\
\multicolumn{1}{c|}{}                                              & \multicolumn{1}{l}{macro}                                         & \multicolumn{1}{l}{micro}                                & \multicolumn{1}{l}{macro}                                & \multicolumn{1}{l}{micro}                                         &                                                                   \\ \hline
MSATT-KG                                                           & 91.0                                                              & \textbf{99.2}                                            & 9.0                                                      & 55.3                                                              & 72.8                                                              \\
MultiResCNN                                                        & 91.0                                                              & 98.6                                                     & 8.5                                                      & 55.2                                                              & 73.4                                                              \\
LAAT                                                               & 91.9                                                              & 98.8                                                     & 9.9                                                      & 57.5                                                              & 73.8                                                              \\
JointLAAT                                                          & 92.1                                                              & 98.8                                                     & 10.7                                                     & 57.5                                                              & 73.5                                                              \\
ISD                                                                & 93.8                                                              & 99.0                                                     & \textbf{11.9}                                            & 55.9                                                              & 74.5                                                              \\
MARN                                                               & 91.3                                                              & 98.8                                                     & 11.6                                                     & 58.4                                                              & 75.4                                                              \\ \hline
PAAT                                                               & \begin{tabular}[c]{@{}c@{}}92.0\\ $\pm$ 0.2\end{tabular}          & \begin{tabular}[c]{@{}c@{}}98.8\\ $\pm$ 0.0\end{tabular} & \begin{tabular}[c]{@{}c@{}}11.1\\ $\pm$ 0.1\end{tabular} & \begin{tabular}[c]{@{}c@{}}59.0\\ $\pm$ 0.2\end{tabular}          & \begin{tabular}[c]{@{}c@{}}75.4\\ $\pm$ 0.1\end{tabular}          \\
\begin{tabular}[c]{@{}l@{}}PAAT\\ +label \\ embedding\end{tabular} & \textbf{\begin{tabular}[c]{@{}c@{}}94.4\\ $\pm$ 0.2\end{tabular}} & \begin{tabular}[c]{@{}c@{}}99.1\\ $\pm$ 0.0\end{tabular} & \begin{tabular}[c]{@{}c@{}}11.4\\ $\pm$ 0.1\end{tabular} & \textbf{\begin{tabular}[c]{@{}c@{}}59.1\\ $\pm$ 0.2\end{tabular}} & \textbf{\begin{tabular}[c]{@{}c@{}}76.0\\ $\pm$ 0.3\end{tabular}} \\ \hline
\end{tabular}
\end{table}

\begin{table*}[!t]
\caption{Comparison results of the PAAT, PAAT without the partition-based encoding (PAAT-PE), PAAT without the partition-based label attention (PAAT-PA), PAAT without both the partition-based encoding and partition-based label attention (PAAT-PEA), and PAAT without bi-LSTM layer (PAAT-BI) on the MIMIC-III 50 (in \%).}
\label{Table3}
\centering
\begin{tabular}{l|rrrrr}
\hline
\multicolumn{1}{c|}{\multirow{2}{*}{Model}} & \multicolumn{2}{c}{AUC}                               & \multicolumn{2}{c}{F1-score}                          & \multicolumn{1}{c}{\multirow{2}{*}{P@5}} \\
\multicolumn{1}{c|}{}                       & \multicolumn{1}{c}{micro} & \multicolumn{1}{c}{macro} & \multicolumn{1}{c}{micro} & \multicolumn{1}{c}{macro} & \multicolumn{1}{c}{}                     \\ \hline
PAAT                                        & \textbf{92.8 ± 0.1}       & \textbf{94.8 ± 0.1}       & \textbf{68.5 ± 0.5}       & \textbf{73.0 ± 0.1}       & \textbf{67.8 ± 0.2}                      \\
PAAT-PA                                     & 92.6 ± 0.1                & 94.7 ± 0.1                & 67.6 ± 0.3                & 72.4 ± 0.2                & 67.5 ± 0.3                               \\
PAAT-PE                                     & 92.7 ± 0.1                & 94.8 ± 0.1                & 67.6 ± 0.4                & 72.6 ± 0.2                & 67.7 ± 0.1                               \\
PAAT-PEA                                    & 92.5 ± 0.2                & 94.7 ± 0.1                & 66.7 ± 0.5                & 72.2 ± 0.2                & 67.5 ± 0.1                               \\
PAAT-BI                                     & 92.4 ± 0.1                & 94.4 ± 0.1                & 67.1 ± 0.3                & 71.8 ± 0.3                & 66.9 ± 0.1                               \\ \hline
\end{tabular}
\end{table*}

\subsection{Baselines}
We evaluated our method with the following baselines.

\subsubsection{MSATT-KG} The Multi-Scale Feature Attention and Structured Knowledge Graph Propagation proposed by Xie et al. \cite{xie2019ehr} uses densely connected CNN to obtain variable $n$-gram features and applies multi-scale feature attention to select features considering the context in different ranges. It also employs the graph CNN to obtain the hierarchical relationships among the ICD codes.
\subsubsection{MultiResCNN} The Multi-Filter Residual Convolutional Neural Network proposed by Li and Yu \cite{li2020icd} applies the multi-filter convolutional layer to capture various text patterns and builds a residual path to enlarge the receptive field of the model.
\subsubsection{LAAT \& JointLAAT} Label Attention Model for ICD coding proposed by Vu et al. \cite{vu2021label} applies a bi-LSTM layer for text encoding and extracts label-specific feature representation using the label attention mechanism. Joint-LAAT is an extension model of LAAT which applies hierarchical joint learning to reflect the hierarchical structure of ICD codes. 
\subsubsection{ISD} Interaction Shared Representation Network with Self-Distillation Mechanism proposed by Zhou et al. \cite{zhou2021automatic} captures the internal connections among codes with different frequencies by applying a self-distillation learning mechanism incorporating code descriptions.
\subsubsection{MARN} Multitask Balanced and Recalibrated Network proposed by Sun et al. \cite{sun2021multitask} captures code associations by utilizing multitask learning using clinical classifications software (CCS) code as additional information. It uses down- and up-sampling based on CNN to handle noisy and lengthy documents and utilizes the focal loss to alleviate the imbalanced class problem.
\subsubsection{HiLAT} Hierarchical Label-wise Attention Transformer Model (HiLAT) proposed by Liu et al. \cite{liu2022hierarchical} divides input text and encodes each segment with ClinicalPlusXLNet. Afterward, the hierarchical label attention mechanism is applied in token level and chunk level. HiLAT did not disclose the MIMIC-III Full dataset result \cite{liu2022hierarchical}. We tried to reproduce it but, we failed due to resource problem. We added the reproduced results on the MIMIC-III 50 dataset.

\begin{figure*}[!t]
\centering
\includegraphics[width=1.8\columnwidth]{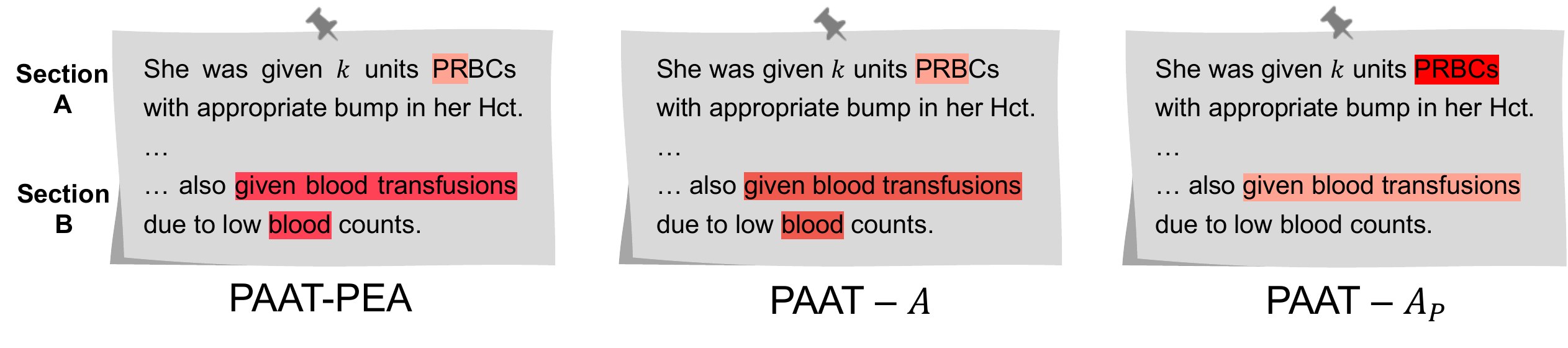}
\caption{The attention maps of the PAAT and PAAT-PEA models for the case of ICD code 99.04 (Transfusion of packed cells). PAAT-$A$ and PAAT-$A_p$ indicates the attention maps of the convolutional and partition-based attention layers, respectively. The highlighting with more intense color indicates the stronger attention.}
\label{Fig.3}
\end{figure*}

\section{Results}
\subsection{Main Results}
Tables \ref{Table1} and \ref{Table2} show the experiment results of the baseline models and our PAAT model on the MIMIC-III 50 and the MIMIC-III Full datasets, respectively. On the MIMIC-III 50 dataset, PAAT showed a similar or higher performance than the baseline models except HiLAT. In particular, the PAAT model achieved higher scores in both macro and micro F1-scores than the baseline models except HiLAT on the MIMIC-III 50 dataset. On the MIMIC-III Full dataset, PAAT achieved the best micro F1-score which is 0.6\% or more higher than the baseline models, but it scored a lower macro F1-score than the ISD and MARN. 
In the training set of the MIMIC-III Full dataset, about 62.3\% of ICD codes have 10 or less training samples. It is difficult to properly train a model only with the training data for those ICD codes. The PAAT model uses no additional information, while ISD uses code descriptions to learn shared representations between codes based on a self-distillation mechanism, and MARN uses additional CCS codes to enhance the capability of capturing the code association. Code descriptions or other medical codes are good candidates for providing additional information to the classification of the ICD codes with the small number of training samples. Therefore, the performance gap in macro results exists between the PAAT model and ISD and MARN. As shown in Table \ref{Table2}, when we applied a label embedding \cite{dong2021explainable} based on the Clinical-Longformer to initialize the label attention layers of the PAAT model, the performance improved in all metrics and achieved similar or higher scores compared to that of the baseline models. In other words, the PAAT model is weak for the case with the small number of training samples, but it can outperform the baseline models when there is additional information for those cases.

Unlike in the MIMIC-III Full dataset, the label embedding did not lead the PAAT model to a noticeable performance improvement on the MIMIC-III 50 dataset. This means that data-based learning is effective enough when the number of training samples is sufficient. In the MIMIC-III Full dataset, when label embedding is applied, the macro F1-score of the PAAT model for ICD codes with 10 or less training samples increased by 0.56\%, while that for ICD codes with more than 10 training samples decreased by 0.11\%. 

\begin{table*}[!t]
\caption{Results when PAAT and PAAT-PEA made different predictions for the MIMIC-III 50 and MIMIC-III Full data sets (in \%).}
\label{Table4}
\centering
\begin{tabular}{c|crcrcrcr}
\hline
\multirow{3}{*}{Model}        & \multicolumn{4}{c}{MIMIC-III Full}                                                                                          & \multicolumn{4}{c}{MIMIC-III 50}                                                                                              \\
                              & \multicolumn{2}{c}{precision}                                & \multicolumn{2}{c}{recall}                                   & \multicolumn{2}{c}{precision}                                 & \multicolumn{2}{c}{recall}                                    \\
                              & macro                            & \multicolumn{1}{c}{micro} & macro                            & \multicolumn{1}{c}{micro} & macro                             & \multicolumn{1}{c}{micro} & macro                             & \multicolumn{1}{c}{micro} \\ \hline
\multicolumn{1}{l|}{PAAT}     & \multicolumn{1}{r}{\textbf{6.1}} & \textbf{34.9}             & \multicolumn{1}{r}{\textbf{8.6}} & \textbf{52.6}             & \multicolumn{1}{r}{\textbf{43.1}} & \textbf{42.1}             & \multicolumn{1}{r}{\textbf{58.9}} & \textbf{59.7}             \\
\multicolumn{1}{l|}{PAAT-PEA} & \multicolumn{1}{r}{5.3}          & 34.4                      & \multicolumn{1}{r}{7.2}          & 47.4                      & \multicolumn{1}{r}{39.1}          & 40.6                      & \multicolumn{1}{r}{40.3}          & 41.1                      \\ \hline
\end{tabular}
\end{table*}

\subsection{Comparison with HiLAT}
HiLAT made such a great contribution to the automatic ICD coding by improving performance based on an efficient model architecture with high performance encoder. However, HiLAT is difficult to apply to the MIMIC-III Full dataset due to the high memory cost because it additionally trains ClinicalPlusXLNet in the training process. For the MIMIC-III 50 dataset, HiLAT requires about 7.1 times of memory resource used in the PAAT model under the same conditions. In our reproduction, when ClinicalPlusXLNet was not trained in the training process, the F1-scores of HiLAT decreased by about 2.4\% in macro and 1.3\% in micro. HiLAT requires resources that are difficult to handle with general devices. The PAAT model recorded slightly lower performance than HiLAT, but is a more efficient ICD coding model.

Inspired by \cite{dong2021explainable}, both HiLAT and the PAAT model segmentally encode the input text and obtain label-specific latent representations for each segment by applying label attention mechanism. And they similarly integrates the latent representations of each segment based on the second label attention mechanism. However, PAAT obtains a global representation based on the bi-LSTM layer for the separately encoded text representations and improves the efficiency of label attention mechanism by using the same $U$ and $W$ in the all label attention process. It also protects important dispersed information from being omitted in softmax operation by applying a smoothing constant $\alpha$. In addition, it enables classification using both the global and local feature representations by allowing the classifier to receive both $V$ and $V_p$ as inputs.

\subsection{Efficiency of the partition-based mechanism}
To understand the effectiveness of the partition-based mechanism, we conducted an ablation study on MIMIC-III 50 dataset. We compared the PAAT model, PAAT without partition-based encoding (PAAT-PE), PAAT without partition-based label attention (PAAT-PA), and PAAT without both partition-based encoding and partition-based label attention (PAAT-PEA). As shown in Table \ref{Table3}, the PAAT model showed the best performance and the PAAT-PEA model recorded the worst performance.

Table \ref{Table4} shows the precision and recall of the PAAT and PAAT-PEA models for the cases where they made different predictions on the MIMIC-III 50 and the MIMIC-III Full datasets. In the MIMIC-III Full dataset, 19,035 cases out of 30,037,776 cases had different decisions, and in the MIMIC-III 50 dataset, 2,331 cases out of 86,450 cases had different decisions. For these cases, the PAAT model achieved higher precision and recall than the PAAT-PEA model. In particular, there is a large gap of more than 10\% of scores in the recall, which means that the PAAT model captures locally important information prone to be ignored by using $V_p$ generated by partition-based label attention. Figure \ref{Fig.3} visualizes the attention maps of the PAAT (PAAT-$A$ and PAAT-$A_p$) and PAAT-PEA models. The PAAT-PEA model missed `PRBCs (packed red blood cells)' containing important information about the ICD code 99.04 (Transfusion of packed cells) corresponding to this case, whereas an attention map PAAT-$A_p$ accurately captured it. Accordingly, the PAAT-PEA model did not discover ICD code 99.04, whereas the PAAT model did.

\begin{table}[]
\caption{The results of the PAAT model according to the number of partitions on the MIMIC-III 50 dataset (in \%).}
\label{Table5}
\begin{tabular}{l|rrrrr}
\hline
\multicolumn{1}{c|}{\multirow{2}{*}{\begin{tabular}[c]{@{}c@{}}Number of\\ Partitions\end{tabular}}} & \multicolumn{2}{c}{AUC}                                                                                                       & \multicolumn{2}{c}{F1-score}                                                                                                  & \multicolumn{1}{c}{\multirow{2}{*}{P@5}}                      \\
\multicolumn{1}{c|}{}                                                                                & \multicolumn{1}{c}{macro}                                     & \multicolumn{1}{c}{micro}                                     & \multicolumn{1}{c}{macro}                                     & \multicolumn{1}{c}{micro}                                     & \multicolumn{1}{c}{}                                          \\ \hline
no-partition                                                                                         & \begin{tabular}[c]{@{}r@{}}92.5\\ ± 0.2\end{tabular}          & \begin{tabular}[c]{@{}r@{}}94.7\\ ± 0.1\end{tabular}          & \begin{tabular}[c]{@{}r@{}}66.7\\ ± 0.5\end{tabular}          & \begin{tabular}[c]{@{}r@{}}72.2\\ ± 0.2\end{tabular}          & \begin{tabular}[c]{@{}r@{}}67.5\\ ± 0.1\end{tabular}          \\
2                                                                                                    & \begin{tabular}[c]{@{}r@{}}92.8\\ ± 0.1\end{tabular}          & \begin{tabular}[c]{@{}r@{}}94.8\\ ± 0.1\end{tabular}          & \begin{tabular}[c]{@{}r@{}}68.0\\ ± 0.6\end{tabular}          & \begin{tabular}[c]{@{}r@{}}72.7\\ ± 0.2\end{tabular}          & \begin{tabular}[c]{@{}r@{}}67.8\\ ± 0.4\end{tabular}          \\
4                                                                                                    & \begin{tabular}[c]{@{}r@{}}92.8\\ ± 0.0\end{tabular}          & \begin{tabular}[c]{@{}r@{}}94.8\\ ± 0.1\end{tabular}          & \begin{tabular}[c]{@{}r@{}}67.8\\ ± 0.4\end{tabular}          & \begin{tabular}[c]{@{}r@{}}72.6\\ ± 0.2\end{tabular}          & \begin{tabular}[c]{@{}r@{}}67.6\\ ± 0.4\end{tabular}          \\
6                                                                                                    & \textbf{\begin{tabular}[c]{@{}r@{}}92.8\\ ± 0.1\end{tabular}} & \textbf{\begin{tabular}[c]{@{}r@{}}94.8\\ ± 0.1\end{tabular}} & \textbf{\begin{tabular}[c]{@{}r@{}}68.5\\ ± 0.5\end{tabular}} & \textbf{\begin{tabular}[c]{@{}r@{}}73.0\\ ± 0.1\end{tabular}} & \textbf{\begin{tabular}[c]{@{}r@{}}67.8\\ ± 0.2\end{tabular}} \\
8                                                                                                    & \begin{tabular}[c]{@{}r@{}}92.7\\ ± 0.1\end{tabular}          & \begin{tabular}[c]{@{}r@{}}94.8\\ ± 0.0\end{tabular}          & \begin{tabular}[c]{@{}r@{}}68.1\\ ± 0.3\end{tabular}          & \begin{tabular}[c]{@{}r@{}}72.7\\ ± 0.2\end{tabular}          & \begin{tabular}[c]{@{}r@{}}67.6\\ ± 0.2\end{tabular}          \\
10                                                                                                   & \begin{tabular}[c]{@{}r@{}}92.8\\ ± 0.1\end{tabular}          & \begin{tabular}[c]{@{}r@{}}94.8\\ ± 0.1\end{tabular}          & \begin{tabular}[c]{@{}r@{}}67.8\\ ± 0.6\end{tabular}          & \begin{tabular}[c]{@{}r@{}}72.7\\ ± 0.1\end{tabular}          & \begin{tabular}[c]{@{}r@{}}67.8\\ ± 0.2\end{tabular}          \\ \hline
\end{tabular}
\end{table}

The discharge summary can be written in long text of more than 10,000 tokens. The partition-based encoding technique is a good way to solve the token limitation problem of transformer encoders that occurs when a long discharge summary enters. The PAAT model can encode up to 4096 tokens per segment by using the Clinical-Longformer, so it can also be applied to a long discharge summary up to 4096$\times n$ tokens. For discharge summaries whose length exceeds 4096 tokens in the MIMIC-III 50 dataset, the PAAT model achieved micro F1-score 1.81\% and macro F1-score 0.41\% higher than the PAAT-PEA model, respectively.

To effectively unify the segmentally encoded feature representations, the PAAT model uses the bi-LSTM layer. As can be seen in Table \ref{Table3}, PAAT without the bi-LSTM layer (PAAT-BI) model showed lower performance in all metrics compared to the PAAT model.

We conducted an additional experiment to find an optimal number of partitions. As can be seen in Table \ref{Table5}, the performance is better when there is a partition. The performance increases as the number of partitions increases up to 6. If the input text is divided too much, the same clinical event may be encoded separately, which causes the discontinuity of information. Accordingly, the performance according to the number of partitions decreased from 6 or more. 

\section{Conclusion}
This paper proposes the partition-based encoding and partition-based label attention for the ICD coding task considering the characteristic of discharge summary. We extract local representations from each section of the input text and combine them with the bi-LSTM layer in the encoding process. Then, with this text representation, we extract two label-specific latent representations. One is obtained by conducting conventional label attention to the latent representation, and the other is obtained by applying the label attention for divided latent representation as in the encoding process. With those two label attention mechanism, our proposed model can effectively capture the important information dispersed in the entire discharge summary which may be missed in the conventional label attention mechanism. The results of our experiments conducted on the MIMIC-III dataset show that our proposed method achieved similar or higher performance compared to the state-of-the-art methods.

% Use \bibliography{yourbibfile} instead or the References section will not appear in your paper
\bibliography{references}

\end{document}